\documentclass[12pt]{article}

\usepackage{a4wide}

\usepackage{amsfonts, amsmath, amssymb}
\usepackage{color}
\usepackage{tikz,url}
\usetikzlibrary{calc,shapes}

\newcommand{\et}{\wedge}
\newcommand{\vel}{\vee}

\renewcommand{\phi}{\varphi}

\newcommand{\dia}[1]{\langle{#1}\rangle}

\newcommand{\M}{\widehat{K}}

\usepackage{newproof}

\newcommand{\weg}[1]{}

\newcommand{\solvable}{\ensuremath{\mathbf{solvable}}}

\usepackage{bbold}

\usepackage{bbold}
\usepackage{bm}

\begin{document}

\title{History of the Muddy Children Puzzle}
\author{Hans van Ditmarsch \\ CNRS, France \ \& \ IIT Kanpur, India \\ {\tt hansvanditmarsch@gmail.com}}
\date{}

\maketitle

\begin{abstract}
The Muddy Children Puzzle is a puzzle about knowledge and ignorance that has been inspiring for the development of epistemic logic. Who came up with it first? This is unclear. We trace the origin of the Muddy Children Puzzle through logical and literary publications over the past two centuries. The puzzle inspired a numerous variations such as involving numbers or coloured hats. We also present a novel hats puzzle involving self-reference.\footnote{This little note is dedicated to the memory of Joe Halpern, who greatly contributed to making the Muddy Children Puzzle popular and a focus of research in epistemic logic, and over many publications, most notably {\em Cheating husbands and other stories} \cite{mosesetal:1986}, co-authored with Yoram Moses and Danny Dolev, and {\em Reasoning about Knowledge} \cite{faginetal:1995}, co-authored with Ronald Fagin, Yoram Moses, and Moshe Vardi. My presentation in this note is non-formal and only precise. Which, of course, was a challenge, as it is much easier to throw formulas around. In this style of presentation Joe Halpern also inspired me, as he was an undisputed master, as again evidenced in \cite{mosesetal:1986}.}$^,$\footnote{I would very much like to know more about history of the Muddy Children Puzzle or related epistemic puzzles before the 1940s, in particular concrete evidence from the 1920s and 1930s, such as involving Alonzo Church. Please write me!}
\end{abstract}

\section{Introduction}
A typical version of the Muddy Children Puzzle is as follows.
\begin{quote} {\em
A group of children has been playing outside and they are called back into the
house by their father. The children gather round him. As one may imagine,
some of them have become dirty from the play. In particular: they may
have mud on their face. Children can only see whether other children
are muddy, and not if there is any mud on their own face. All this is
commonly known, and the children are, obviously, perfect logicians. Father
now says: ``At least one of you is muddy.'' And then:
``Will those who know whether they are muddy step forward.'' If
nobody steps forward, father keeps repeating the request. At some stage 
all muddy children will step forward. When will this happen if $m$ out of $k$ children in total are muddy, and why?
} \hfill \cite{hvdetal.puzzle:2015}
\end{quote}

As is well-known, after father makes his request $m$ times the $m$ muddy children step forward, and at request $m+1$ the $k-m$ (if any) clean children step forward. The round wherein the clean children step forward is often omitted from the analysis, and this also depends on the formulation of the puzzle: if the request is to step forward if you know {\em that} you are muddy, only the $m$ muddy children step forward in round $m$.

We can prove why this answer is correct by induction. The base of the induction is $1$, not $0$, as there is at least one muddy child. Initially, it is uncertain whether all children are clean or it is the only muddy child. After father's annoucement ``At least one of you is muddy'' this uncertainty is resolved, because it sees no muddy children and therefore it will step forward in round $1$. Everyone else will step forward in round $2$. Let us now assume that $k$ muddy children will step forward in round $k$, and let there be $k+1$ muddy children. A muddy child now sees $k$ muddy children and is uncertain if there are only $k$ muddy children, in which case it is clean, or $k+1$ muddy children, in which case it is dirty. As in round $k$ nobody stepped forward, this uncertainty is resolved and it must therefore be muddy and will step forward in round $k+1$. All the $k+1$ muddy children will therefore step forward. The clean children will not step forward, as they are uncertain between $k+1$ and $k+2$ muddy children. But once the muddy children stepped forward, that uncertainty is also resolved and they step forward in round $k+2$.

\section{History of the muddy children puzzle}

As logic puzzles are not considered science, their provenance is often not given. And as all nations have their own tradition in mathematical entertainment and puzzlebooks, they have emerged in many other languages than English, and often crossed linguistic barriers during their dissemination. We have come across various sources for the Muddy Children Puzzle and related puzzles over the past two centuries --- with a curious gap from the 1830s to the 1930s that we have not been able to fill.

\subsection{Before $\bm{1900}$}

\begin{figure}[h] \center
 \includegraphics[width=12cm]{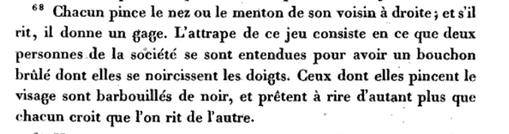}
\begin{quote}
{\em 
Everyone pinches the nose or the chin of his right neighbour. If he laughs, he gives a pledge. The catch of this game is that two people from the company agreed to have a charred piece of cork to blacken their fingers on. Those whose faces they pinch will be smeared with black, and thus inclined to laugh even more as each believes to be laughing about the other one.} \hfill \cite{rabelais:1823}
\end{quote}
\caption{Facsimile of note $68$ on page $413$ of the first volume of the $1823$ edition of the works of Rabelais \cite{rabelais:1823}, with English translation by me}
\label{fig.rabelais}
\end{figure}

Where you want to begin their history all depends on what you still call a puzzle, A fair claim of the oldest source for the muddy children puzzle is an $1823$ annotated version \cite{rabelais:1823} of the 16th century French literary classic `Gargantua et Pantagruel' by Fran\c{c}ois Rabelais \cite{rabelais:1532,rabelais:1534}. Gargantua is a big guy eating a lot and having a lot of fun such as playing a lot of games, expecting to be entertained by his nervous Parisian hosts who do not want to be eaten instead. The editors Charles Esmangart and \'Eloi Johanneau add extensive footnotes to this $1823$ edition, explaining and describing among many other games (such as Trictrac) the game {\em je te pince sans rire} (I pinch you without laughing), see Figure~\ref{fig.rabelais}.\footnote{I am unaware of older editions with the same footnote. I consulted only some and not all older editions. The $16$th centery original also mentions games, but not  `pince-sans-rire', and does not have footnotes.}

The game Pince-sans-Rire is is a somewhat older version of the game Barbichette. Both involve pinching the nose or chin or the person you are facing, or pulling his beard (in case it concerns men). This is of course absolutely ridiculous! And that's the point: the one who cannot resist laughing loses the game from the one who keeps a straight face. (It makes one wonder if there is a relation with the British `stiff upper lip'. Maybe the result of generations of English and French playing Barbichette.) In the Pince-sans-Rire game there is a group of players from which two face each other in each round of the game, and anyone from the group laughing first will be the next victim undergoing pinching. Whereas Barbichette concerns only two players facing each other and pinching each other in turn.\footnote{In the first volume of the Asterix comics series, the protagonist Asterix is playing Barbichette with Roman soldiers \cite[p.\ 15 \& 41]{asterix:1961} (\url{https://archive.org/details/asterix-t-01-asterix-le-gaulois/page/n15/mode/2up}). Roman soldiers shave, unlike cool Gallic tribesmen. First, on page 15, Asterix pulls off the false beard of a spy, posing as a Gaul, thus revealing him to be Roman, and later, on page 41, Asterix now confidently expects a Roman soldier's beard to be false and to come off again. Only this Roman's beard was real and had grown instantly after drinking a magic potion...}
%

The $1823$ French version was found by Hans Rott, when I discussed my historical findings with him. With a little help, as I knew of a somewhat later $1832$ German translation of \cite{rabelais:1823} by Gottlob Regis \cite{regis:1832}. It features the same footnote, in German, and the game is then called `{\em ungelacht pfetz ich dich}', a literal translation (despite `ungelacht' not found in German dictionaries). It was presented and discussed by Hurkens, Born and Woeginger in \cite{EATCS:2008}. When I asked the late Gerhard Woeginger already quite some years ago how on earth he found this source, he stoically responded ``Just type {\em ungelacht pfetz ich dich} in Google and it's the top hit coming up in the search''.

There is clearly a relation between the chins or noses smeared with charred cork in Pince-sans-Rire and muddy foreheads in the Muddy Children Puzzle: you cannot see either, but you can see it on the face of others. But it is not so clear how the older sources are not merely games but also epistemic puzzles. One has to read a bit into the text as published, while following the rules of the Pince-sans-Rire and Barbichette games. If I smear your face and you look more ridiculous with a smeared face, then I'm more inclined to laugh. So I lose the game. At least in Barbichette. But wouldn't I like to win? Pince-sans-Rire seems then a more suitable setting, wherein, say, Alice faces a neighbour Bob blackening her nose while pinching it, in the presence of the remaining onlooking players including Cath laughing their heads off. In case Cath had been blackened in a previous round, the realization may dawn upon her then and there, that this is why Alice was laughing in that previous round, causing Alice to be pinched now. I must have a blackened nose too, concludes Cath! One can also imagine two players facing each other, both blackening the other's face and then laughing simultaneously, which would be a draw! The ``each believes to be laughing about the other one'' suggests such an interpretation. But this contradicts somewhat the prior ``two people from the company agreed to have a charred piece of cork to blacken their fingers on'', because then these two knew in advance what would happen, and the surprise in laughing simultaneously about each other is therefore lost. We are surely reading too much into the literal text... Unlike muddy children, $16$th century players of Pince-sans-Rire are not perfect logicians.

Mud on foreheads or charcoal on noses or chins does not seem to be a big deal. One could imagine a gradual transition towards a formulation more as in epistemic puzzles, throughout the remainder of the 19th and the 20th century, trickling through the literature and puzzle sections of magazines, or by way of oral history. Wouldn't this be a typical Lewis Carroll riddle? It is not. I checked. Would it not be a typical brain teaser in one of the many puzzle collections appearing round the end of the 19th century? I consulted dozens, in all languages I master, and did not find it. (The extensive literature section in the NOB puzzle collection of the JAIST Gallery in Kanazawa, Japan, was a great help.) Such puzzlebooks contain heaps of chess problems, instead! I did not find any source resembling the Muddy Children Puzzle between the 1830s and the 1930s. Pince-sans-Rire and Barbichette continued to be played throughout this time, until recently (and the event of mobile phones and social media). I have a colleague who still played Barbichette when young. Come to think of it, I saw some guys playing it on a packed Toulouse-bound train recently, slapping each other's face while laughing. A close approximation. They had lots of fun. I can therefore imagine some 20th century researcher or problem solver (such as Gerhard Woeginger) to see the connection, and it'd be hard to find an older reference explaining Pince-sans-Rire in this way of an even earlier date than $1823$.

\subsection{After $\bm{1900}$}

The Muddy Children Puzzle seems to make its (re)appearance from the $1930$s onward, simultaneously with or maybe slightly later than (recorded) occurrences of other epistemic riddles, e.g., those by Williams and Savage on ignorance about ages or house numbers in \emph{Perplexities}, their puzzle section in Strand Magazine, published separately in $1935$ \cite{Williams35} and later, in $1940$, in their {\em
Penguin problems book} \cite{Williams40}. A detailed discussion of these puzzles is found in \cite{hvdetal.jlc:2007}. An unconfirmed rumour is that Alonzo Church wrote on the Muddy Children Puzzle in the early $1930$s. Which I could not trace in the Church archive maintained by Princeton University, nor anywhere else. 

The currently earliest 20th century source known to me is Japanese. It came to me by way of Katsuhiko Sano. The physicist Paul Dirac introduced the Muddy Children Puzzle to a Japanese audience during his visit to Japan in $1929$\footnote{The year $1938$ has also been suggested, by Katsuhiko Sano. I could not confirm this later visit from his biography, wherein only the $1929$ visit is mentioned, together with Werner Heisenberg --- apart from a much later one in the $1950$s. An earlier visit would be more intriguing, as even $1938$ would be an early $20$th century indirect reference for such epistemic puzzles.}, which is indirectly confirmed because the Muddy Children Puzzle is known in Japan as {\em Dirac's Riddle}. There is also direct evidence: mystery writer Takataro Kigi wrote a detective novel {\em Window with a view of the sea} motivated by Dirac's Riddle that was published in $1941$ \cite{kigi:1941}. 

\begin{figure}[h] \center
\includegraphics[width=5cm]{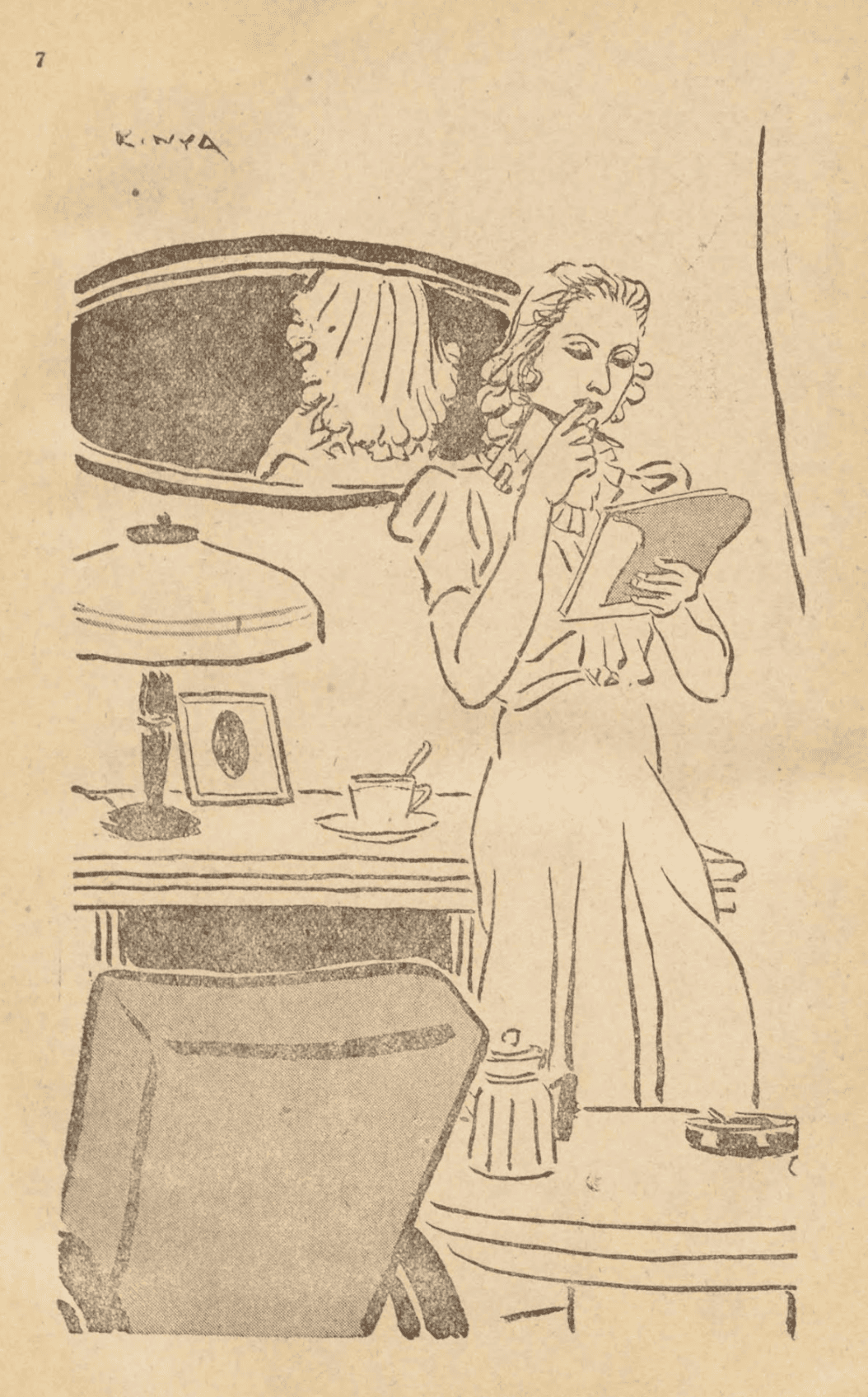}
\caption{Illustration from {\em Window with a view of the sea} featuring Dirac's Problem \cite[p.\ 7]{kigi:1941}}
\label{fig.kigi}
\end{figure}



\begin{figure}[h] \center
\includegraphics[width=9cm]{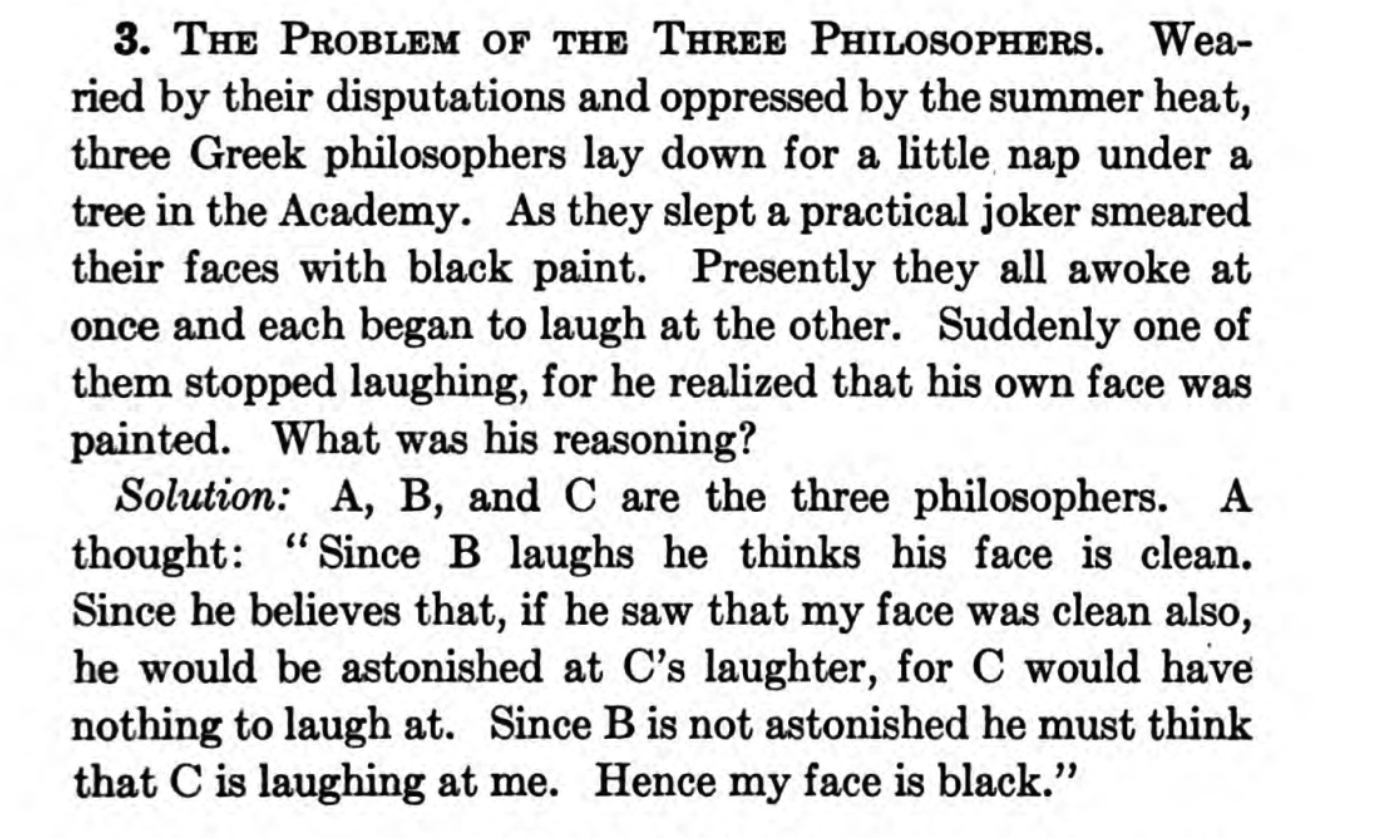}
\caption{The Muddy Children Puzzle in an incarnation for three philosophers. Chapter 1, Mathematics without Numbers \cite[p.\ 15]{kraitchik:1942}}
\label{fig.kraitchik}
\end{figure}

A Muddy Children Puzzle motivated detective novel is still somewhat indirect evidence. On the heals of its publication, the earliest $20$th century direct reference that we know of is from $1942$ and was published in the {\em Mathematical Recreations} by Kraitchik \cite[p.\ 15]{kraitchik:1942}, see Figure~\ref{fig.kraitchik}. Here it should be noted that we also consulted the earlier $1930$ French edition of this puzzlebook \cite{kraitchik:1930}. The Muddy Children Puzzle does not occur in that earlier edition. The career of this (Jewish) mathematician Maurice Kraitchik who grew up in Minsk in Bielorussia reads like that of a continuous refugee. As a Jew not being allowed to study in Russia 
he went to Belgium for his academic education. After building up a career there, and by then publishing in French, he left to the USA at the start of the $2$nd World War, and from that moment onward published in English. The well-known {\em two-envelope problem} also first appeared in {\em Mathematical Recreations} \cite{kraitchik:1942}.

Another fairly early mid $20$th century original version of the Muddy Children Puzzle, for three ladies with dirty faces and very similar to Kraitchik's version, appeared in Littlewood's `{\em A Mathematician's Miscellany}' \cite{littlewood:1953}, right at the beginning.  (This book is also an original source for the {\em consecutive numbers riddle}.)  It is as follows: \begin{quote} {\em Three ladies, $A$, $B$, $C$ in a railway carriage all have dirty faces and are all laughing. It suddenly flashes on $A$: why doesn't $B$ realize $C$ is laughing at her? --- Heavens, I must be laughable. (Formally: if I, $A$, am not laughable, $B$ will be arguing: if I, $B$, am not laughable, $C$ has nothing to laugh at. Since $B$ does not so argue, I, $A$, must be laughable.)}  \hfill \cite[p.\ 3--4]{littlewood:1953} \end{quote} Interestingly enough, Littlewood calls (solving) the puzzle a typical example of non-trivial mathematical reasoning. When discussing the solution he generalizes to the case where all of $n$ children are muddy and proves this by natural induction. The puzzle is a bit of tongue-in-cheek to his colleague Hardy, who in \cite{hardy:1940} cites Euclid's proof that there are infinitely many primes as a typical example of non-trivial mathematical reasoning; a classical mathematical problem, not a modern one. Whereas Littlewood puts Muddy Children in the top spot of `genuine mathematics', opposing it explicitly to Hardy's infinity of primes.

Kraitchik and Littlewood reason by counterfactuals, `if I, $A$, am not laughable, $B$ will be arguing: if I, $B$, am not laughable, \dots'. It seems a small step from there to considering possible worlds and epistemic modal logic as in the later versions by Halpern and co-authors: given actual world $w$, $A$ considers a world $v$ possible wherein she is not laughable, and given $v$, $B$ considers a world $u$ possible wherein he is not laughable, \dots'. But of course there is still a big difference. The Kraitchik and Littlewood versions determine a single action from the perspective of $A$, while $B$ and $C$, although perfect logicians, do not act. But why shouldn't they act the same as $A$? In what order should $A$, $B$ and $C$ then stop laughing? Surely we cannot assume this is simultaneously. In these older versions synchronization or a specific order of different actions is not enforced. With Kraitchik and Littlewood it is a riddle {\em about knowledge}, but with Halpern \& co.\ it is a riddle {\em  about knowledge and action}.

An often quoted $1958$ source is a puzzlebook by Gamow and Stern \cite{gamowetal:1958}. This sketches a situation in the country of Quasiababia involving married men and forty unfaithful wives they are married to. In this version everyone knows if a woman is unfaithful except her own husband. We may assume there are more than forty married couples, so this is the first version proving something for $m$ out of $k$ agents, where $m=40$ and $k$ is an unknown larger quantity (where Kraitchik and Littlewood treat the case where $m=k=3$, and the Pince-sans-Rire original goes for $m=k=2$).

From the $1950$s onward the Muddy Children Puzzle is clearly established as mathematical entertainment folklore in many publications, of which we name a few below. I find it remarkable that different versions of the riddle also reflect morality (should it be about sexual mores, why not about coloured hats, or mud) and attempts to be more inclusive towards the other half of humanity; where we recall that the $1950$s is also the era wherein professional orchestras started to do blind auditions --- from which time women became regular orchestra players, unlike orchestra conductors who were not auditioned in that way.

In \cite{mccarthy:1978} (based on a late $1970$s manuscript) John McCarthy publishes an non-abusive version for `wise men'. The wise men have to find out whether the dot (`spot') on their forehead is black or white. Versions with hats in two colours also pop up around then, and even earlier as in \cite{vantilburg:1956} presented in the subsection on Hat Puzzles, in which puzzles for more than three colours are also discussed.

Gardner has various publications on epistemic and other puzzles from the $1970$s onward in journals such as the {\em Scientific American} and the {\em Mathematical Intelligencer}, and also in e.g.\ {\em Isaac Asimov's Science Fiction Magazine}. A collection of Gardner's puzzles in the last was bundled in \cite{gardner:1984}, wherein he presents a version of Muddy Children for women suffering unfaithful husbands, with the giveaway title `the castrati of Womensea' --- Womensa is the planet providing the riddle's setting; that must have been the connection to science fiction.
 
Moses, Dolev, and Halpern also present a version with unfaithful men, in \cite{mosesetal:1986}, now enacting a dynasty of Henriettas in the town of Mamajorca on the sunken island of Atlantis (the irate women now shoot their unfaithful husbands instead of castrating them). A slightly earlier version \cite{halpernmoses:85b} by Halpern and Moses presents the same version but with muddy children, based on and similar to \cite[p.\ 382]{barwise:saos} (Barwise). Then, in {\em Reasoning about Knowledge} \cite{faginetal:1995}, they and their co-authors weave the entire first two chapters containing the basics of epistemic logic around the Muddy Children Puzzle. What I find remarkable in their formulation is father's request is:
\begin{quote} {\em Does any of you know whether you have mud on your own forehead?} \hfill \cite[p.\ 4]{faginetal:1995}
\end{quote}
That is, `whether', and not `that' as in their prior \cite{halpernmoses:85b} incarnation of muddy children, in \cite{barwise:saos} they quote, and common in many other venues for the puzzle. The `whether' version seems to require all clean children to step forward after all dirty children. Possibly, that next round was considered too obvious. Hat puzzles, discussed below, may involve even more than two rounds wherein agents act (and more than two colours), so that diverging behaviour over different rounds is not obvious.

Note the straight correspondence from the `faces smeared with black' in Rabelais via the `faces smeared with black paint' in Kraitchik to the `muddy faces' in Halpern \& co. It can be coincidence. But it suggests hidden links that I am unaware of --- but would love to be told about by readers.

\begin{figure}[h] \center
 \includegraphics[width=6.5cm]{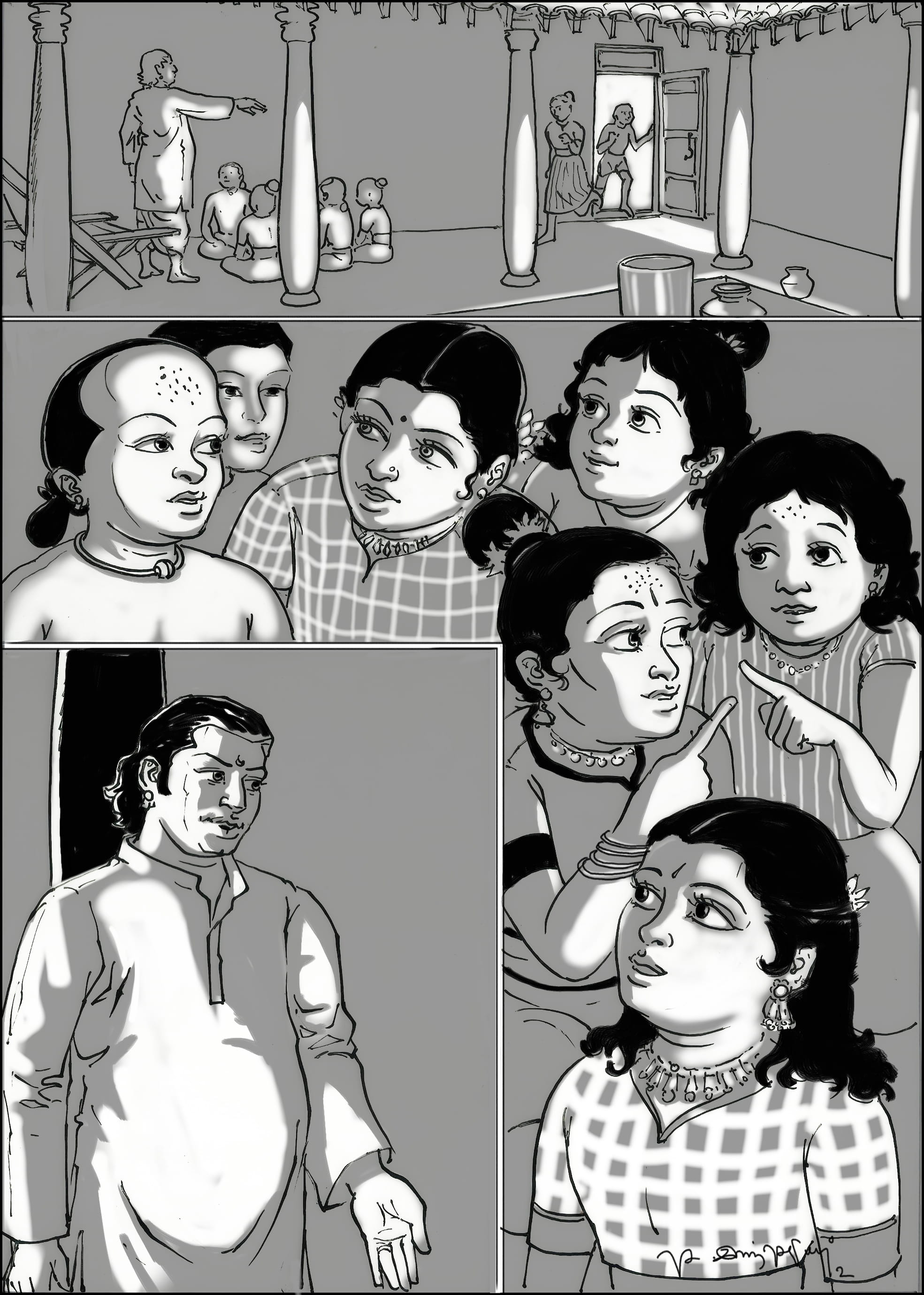}
\caption{Muddy Children in an Indian setting, by Elancheziyan \cite[p.\ 22]{hvdetal.puzzle:2015}}
\label{fig.india}
\end{figure}

\paragraph*{From epistemic puzzles to epistemic logic} From around $1980$ onward the further dissemination of the riddle was intertwined with the development of epistemic logic (the modal logic of knowledge), artificial intelligence (then seen as the formalization of the knowledge and actions of human and artificial agents), and computer science (communicating processes and distributed computing). McCarthy \cite{mccarthy:1978} gives a first-order logical analysis (using the standard translation of modal logic into first-order logic), Barwise \cite{barwise:saos} analyzes such puzzles as evolving  \emph{situations} involving perception (that implies knowledge), and only Moses, Halpern, and co-authors \cite{halpernmoses:85b,faginetal:1995} firmly link the subject to multi-agent epistemic logic and its dynamic extensions. (Their \cite{mosesetal:1986} is non-formal --- but no less precise, and it focuses on versions involving different kinds of asynchrony, and error.) \emph{That} firmly, one could say, that the muddy children version from \cite{faginetal:1995} became `de rigueur' in the scientific literature on multi-agent epistemic logic. There are some diverging approaches for the dynamics of such epistemic logical semantics. Apart from the runs-and-systems approach in \cite{faginetal:1995} where the information changing events (such as announcements) are described as knowledge-based programs, and where their execution corresponds to a point in time so that the resulting logics are {\em temporal epistemic logics}, a different approach where such events  correspond to modalities that \emph{update} models also became quite influential. The starting publication here is by Plaza \cite{plaza:1989} and the subsequent development of so-called {\em dynamic epistemic logics} is evidenced by publications such as \cite{baltagetal:1998,hvdetal.del:2007}. A recent puzzlebook treating Muddy Children and other epistemic riddles in that way is \cite{hvdetal.puzzle:2015}, from which Figure~\ref{fig.india} displays Muddy Children in an Indian setting.

\section{Fifty shades of mud}

In this section we present by example further variations of the Muddy Children Puzzle in the subsequent mathematical entertainment literature, and some cultural connotations and reflections.

\subsection{The art of indirect communication}

Muddy Children, again; consider Alice, Bob and Cath in the presence of their father who was calling them because it's dinner time. It seems a great deal more confrontational when in the presence of Bob and Cath father points at Alice while saying `you are muddy, go wash your face!'. How embarrassing! And much more indirect if he says `one of you is muddy, if you know that you're muddy go wash your hands'. They start by looking at each other, and eventually, without any fingers being pointed, Alice and the other muddy ones from those three, if any, eventually go wash their face. 

Partially this is about the difference between `everyone knows' and `common knowledge', and about who dares to bridge the gap, as discussed at length in books on the societal and psychological aspects of common knowledge \cite{chwe:2001,pinker:2025}. But partially it seems also to be delayed execution (exposure \dots) of some kind: after initially creating \emph{some} common knowledge however without targetting an individual (`at least one of you is muddy'), father repeats the request to step forward and go wash your face, in case it concerns you, multiple times, after which eventually, again without pointing fingers, the desired outcome appears: common knowledge who is muddy, but in a much gentler, delayed, fashion. The culprits themselves figure it out. No pointing of fingers.

This feature seems common in all (unsanitized) versions of the riddle, and even in the original Pince-sans-Rire: somewhat you don't want to be the ridiculous one; but if other people also laugh at you, it becomes unavoidable. You do not want your friend to tell you `your wife is unfaithful' or `your husband is unfaithful'. Far too direct. In India there is a version for blue and brown eyes (personal communication by an Indian collaborator). You cannot see the colour of your own eyes. Stating to an Indian person that he or she has blue eyes is like stating that their parent or ancestor is not Indian but English, so maybe shameful (again, `unfaithful' crops up, forced or otherwise). 

Then again, maybe this is cheap talk. It can just be the cultural setting needed for puzzlebooks or good stories, where the factual information should be juicy or embarrassing, to get more readers. Guaranteed embarrassment if it's about sex. Why would you otherwise not give the factual information directly? 

In hat puzzles, discussed next, the aspect of avoiding social embarrassment has disappeared completely. Well, there's the hats. I would personally still find it somewhat embarrassing to walk around town in a green or yellow floppy hat, walking straight out of Disney's Snow White.

\subsection{Hat Puzzles}

Just as you cannot see mud on your own forehead, you cannot see the colour of a hat you're wearing. Such hat puzzles also go back to, at least, the $1950$s.

A $1956$ hat puzzle with sequential announcements is found in the weekly magazine \emph{Katholieke Illustratie} \cite{vantilburg:1956}. This is one of $626$ such ``breinbrouwsels'' (brain brews) that appeared in that magazine between $1954$ and $1965$, namely Breinbrouwsel $137$ (volume $90$, issue $32$, page $47$) entitled `Doe wel en zie niet om' (Do well and don't look back). There are three red and four blue caps, worn by a Dutch Olympic team\footnote{A version with the three colours red, white, and blue of the Dutch flag was presumably too complex.} of four rowers sitting in front of one another, where a rower can only see the colour of the caps in front of him, but not the colour of his own cap or that of those behind him. The puzzle is played like a repeated game, namely over several days of rowing, where each new day and next race each rower gets a randomly chosen hat colour put on his head by the coach (so you cannot see the colour). On the first day, the backmost rower says that he knows the colour of his hat. Therefore \dots? (Please answer!) And so on. Sometime in the middle of the first decade of this millennium Rineke Verbrugge and I spent several days in the Groningen University Library leafing through decades of volumes of this magazine \emph{Katholieke Illustratie}, uncovering this information \cite{hvdetal.jlc:2007}.\footnote{By now I could check details again in an online version of all issues of that weekly magazine, from the late $19$th century onward. It was much more fun doing it with Rineke, leafing through actual volumes in a real library. At the time we were happy to have uncovered an older source for a Muddy-Children-like riddle than the $1958$ one by Gamow and Stern \cite{gamowetal:1958} already discussed above, that we then thought to be the oldest on record. How much mistaken we were! I only discovered Littlewood from $1953$ and Kraitchik from $1942$ later, and only after that the much older $19$th century sources.} 

\begin{figure}[h] \center
 \includegraphics[width=7.5cm]{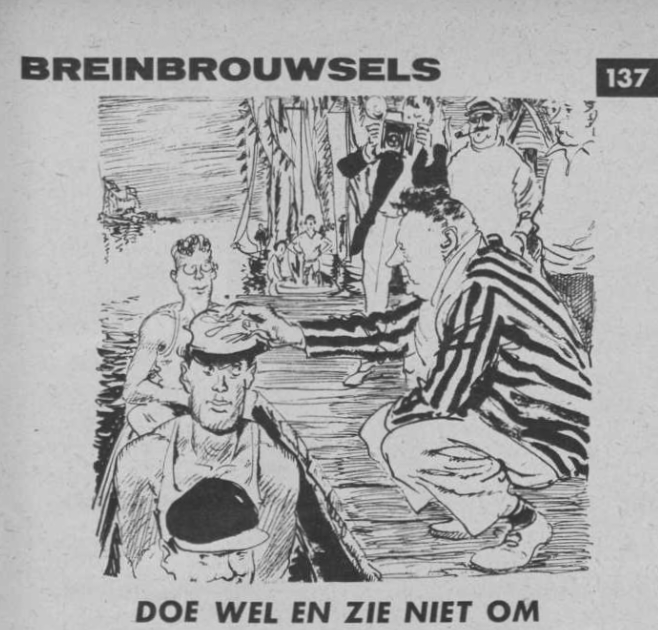}
\caption{The Dutch rowing team in action, as a hat puzzle \cite{vantilburg:1956}}
\label{fig.vantilborg}
\end{figure}

We cannot fail to also quote a not so different by slightly more general (three instead of two colours, and questions about ruled out colours) $1982$ hat puzzle from Smullyan's well-known puzzlebook `{\em The Princess and the Tiger}' \cite{smullyan:1982}.

\begin{quote}
{\em Three subjects---$A$, $B$, and $C$---were all perfect logicians. Each could instantly deduce all consequences of any set of premises. Also, each was aware that each of the others was a perfect logician. The three were shown seven stamps: two red ones, two yellow ones, and three green ones. They were then blindfolded, and a stamp was pasted on each of their foreheads; the remaining four stamps were placed in a drawer. When the blindfolds were removed, $A$ was asked, ``Do you know one color that you definitely do not have?'' $A$ replied, ``No.'' Then $B$ was asked the same question and replied, ``No.''

Is it possible, from this information, to deduce the color of $A$'s stamp, or of $B$'s, or of $C$'s?} \hfill \cite[p.\ 6--7]{smullyan:1982}
\end{quote}

After the $1980$s, Hat Puzzles developed independently from Muddy Children Puzzles, not analyzed in epistemic modal logic but in mathematical statistics, for example in work by Hardin and Taylor in the {\em Mathematical Intelligencer} \cite{hardinetal:2008} and in their book  \cite{hardinetal:2013}  --- and these are just two of many publications in that tradition. That puzzle community focused on game and combinatorial aspects, including countably (infinite or co-finite) numbers of hats, or even uncountably many. The aim there is to maximize the probability for agents to guess the colour of their hat correctly. In the publications in epistemic logic the focus is on knowledge, that is, probability $1$. It seems remarkable that these statistical and logical traditions did not overlap over the past decades. 

Remarkable, because there also much more involved epistemic hat puzzles. To such a hat puzzle, a high point in puzzle math that entirely stands on its own, we dedicate a separate later Section~\ref{section.muetzen}.

\subsection{Consecutive Numbers and Sum and Product}

Instead of Muddy Children with muddy or clean faces, or hat puzzles with hats in one of many different colours, there are also many epistemic riddles involving uncertainty about natural numbers. Now this could be $0$ for clean and $1$ for muddy, but also for any finite set of natural numbers (such as ages and house numbers in the $1930$s and $1940$s \cite{Williams35,Williams40}), and even for all infinitely many natural numbers, such as the well-known {\em Consecutive Number Puzzle}, of which the first known occurrence is as follows.
\begin{quote}
{\em There is an indefinite supply of cards marked $1$ and $2$ on opposite sides, and of cards marked $2$ and $3$, $3$ and $4$, and so on. A card is drawn at random by a referee and held between the players $A$, $B$ so that each sees one side only. Either player may veto the round, but if it is played the player seeing the higher number wins. The point now is that every round is vetoed. If $A$ sees a $1$ the other side is $2$ and he must veto. If he sees a $2$ the other side is $1$ or $3$; if $1$ then $B$ must veto; if he does not then $A$ must. And so on by induction.}  \hfill \cite[p.\ 4]{littlewood:1953} \end{quote} 
We recall Littlewood as a source of Muddy Children, on page $3$ of the same, followed on the next page by Consecutive Numbers, as another example of `modern math'. In the Littlewood version of Consecutive Numbers there is no `solution' (every round is vetoed) and the synchronization is left open to interpretation (who vetoes first?). Also note that he implicitly lets the natural numbers start with $1$, not $0$, which sometimes leads to confusion in similar riddles (and in particular across language barriers).
Again there are many other versions, such as Gardner's \cite{gardner:1979}, and, fairly typical:
\begin{quote} {\it
Anne and Bill get to hear the following: ``Given are two natural numbers. They are consecutive numbers. I am going to whisper one of these numbers to Anne and the other number to Bill.'' This happens. Anne and Bill now have the following conversation.
\begin{itemize}
\item Anne: ``I don't know your number.''
\item Bill: ``I don't know your number.''
\item Anne: ``I know your number.''
\item Bill: ``I know your number.''
\end{itemize}
First they don't know the numbers, and then they do. How is that possible? What surely is one of the two numbers?
} \hfill \cite{hvdetal.puzzle:2015} \end{quote}
Consecutive Numbers is also known as the {\em Conway Paradox}, after its occurrence in \cite{Emde:84} by Van Emde Boas, Groenendijk and Stokhof, that was influential in the development of dynamic epistemic logic by way of its roots in {\em update semantics} \cite{veltman:1996}. This attribution to Conway was mistaken, as Peter van Emde Boas could ascertain us many years later. {\em A Headache-Causing Problem} by Conway and Patterson \cite{conwayetal:1977} also involves uncertainty about numbers but, instead, uncertainty about their sums. (See the detailed discussion in the historical notes to \cite[Chapter 1]{hvdetal.puzzle:2015}.)

Talking about sums, and products, in this list we should not fail to mention Freudenthal's $1969$ {\em Sum and Product} riddle \cite{freudenthal:1969}, that made the rounds of multiple of Gardner's puzzle columns in the Scientific American in the late $1970$s \cite{gardner:1979}, and that was much later (with many variations) presented in exquisite historical detail in \cite{EATCS:2006,EATCS:2007,EATCS:2008} as well as in \cite{hvdetal.jlc:2007,hvdetal.puzzle:2015}. There is a curious several years gap in how it crossed the Atlantic from the Netherlands to the USA, where McCarthy picked it up in the late $1970$s \cite{mccarthy:1978}.

\begin{quote} {\em
$A$ says to $S$ and $P$: I have chosen two integers $x, y$ with $1 < x < y$ and $x+y \leq 100$. In a moment I will inform $S$ of their sum $s = x+y$, and I will inform $P$ of their product $p = xy$. These announcements will remain secret. You are required to make an effort to determine the numbers $x$ and $y$.

He does as announced. The following conversation now takes place:
\begin{enumerate} 
\item $P$ says: I don't know the numbers. 
\item $S$ says: I knew you didn't know the numbers. 
\item $P$ says: Now I know the numbers. 
\item $S$ says: Now I also know the numbers. 
\end{enumerate} 
Determine $x$ and $y$.
} \hfill \cite[translation from Dutch]{freudenthal:1969}
\end{quote}
I find this riddle a high point in epistemic puzzles, because it involves \emph{second-order} epistemic uncertainty: to solve the riddle it is crucial that Sum knows that Product does know the numbers.

\paragraph*{Non-public dynamics}

With numbers on foreheads, or being whispered into agents' ears, we go straight from two to many (Sum and Product) or infinitely many (Consecutive Numbers) possibilities (or transfinitely many, as in \cite{parikh:1992} discussed below), and engage into `epistemic number theory'. With colours of hats we seem to have the same options, although it is harder to imagine riddles with infinitely many colours, because how should one distinguish them by observation? One then goes rather into other kinds of riddles, such as the {\em Sorites Paradox} \cite{sep-sorites-paradox} and other epistemic probabilistic riddles. Yet other epistemic puzzles involve self-reference between the announcements and the subsequent epistemic desiderata, such as the {\em Surprise Examination} \cite{HalpernM86,hvdetal.synthese:2006,gerbrandy:2007}. In all epistemic puzzles entering the scene so far the dynamics is public. All agents simultaneously incorporate the new information, what is known as {\em full synchronization}. There are also epistemic puzzles with non-public dynamics, such as {\em One Hundred Prisoners and a Light Bulb} \cite{dehayeetal:2004,hvdetal.puzzle:2015} and the {\em Gossip Problem} \cite{BAKER1972191}. And it seems time to take another, closer, look at the dynasty of Henriettas featuring in \cite{mosesetal:1986}.

\subsection{Asynchrony, lying, cleaning, and infinity}

We finish this section with a number of straight variations on Muddy Children that not so much inspired the puzzle community but rather further research in epistemic logics, in fact, were also seen as embodying such research. We do not wish to be exhaustive here, and, with the exception of the last variation, merely illustrate these variations for three children Anne, Bill, and Cath.

\paragraph*{Asynchrony}

We first recall the disgraceful reign of Henrietta II in \cite{mosesetal:1986}. Instead of publicly announcing that there is at least one unfaithful husband, and the protocol to be followed for wives shooting their husbands, she informs her subjects of that information in individual letters that are known to eventually arrive. But not on which day \dots \ A slightly adapted equivalent for three muddy children Anne, Bill and Cath (much less realistic than the \cite{mosesetal:1986} setting and also a great deal more boring) is that father's first announcement is not ``At least one of you is muddy''  but ``I will privately give each of you an envelope containing factual information on muddiness'' (and we may assume this is truthful and that these envelopes will in due time actually be given --- the children trust their father), following which the riddle proceeds as before. 

\begin{quote} {\em 
In fact, all of Anne, Bill and Cath received the envelope before father's next request to step forward if you know whether you are muddy.

Let us first assume Anne is muddy. She will step forward, as she received the envelope and sees that Bill and Cath are clean. 

Let us now assume that Anne and Bill are muddy. At the first request, nobody steps forward. Anne considers it possible that all received the information, but even so she would only expect to step forward at the next request. However, at the second request she is in doubt. Suppose Bill has not received the envelope, then he does not know that at least one child is muddy, so even if I, Alice, were clean, he would have had no reason to step forward at the first request. So I'd better not act on him having received that information, and will not step forward now. And neither does Bill. Nor Cath, for other reasons. Nor will anyone ever step forward, no matter how often father repeats his request.}
\end{quote}

Variations between full synchrony and full asynchrony are also discussed in \cite{mosesetal:1986}, where we recommend to read up on the further history of Mamajorca involving the reign of Henrietta III. All this is concurrent with and predates many subsequent works in asynchronous, or partially synchronized, distributed systems and their dynamic development (over runs of events and actions) \cite{halpernmoses:1990}, and later work in dynamic epistemic logics on asynchronous announcements \cite{BalbianiDG22}.

\paragraph*{Lying and error}

Further complications in \cite{mosesetal:1986}, during the reign of the next Henrietta IV, involve `disobedient wives' that do not shoot their unfaithful husbands even when they know that they are unfaithful (``please promise me not to do it again''), where however common knowledge that at least one wife is obedient (and acts correctly according to the protocol: shooting the guy) still allows for all obedient wives to eventually shoot their husbands. This clearly relates to {\em protocols with errors} and involving {\em byzantine agents} in distributed computing, and the many results obtained for the behaviour of $n-f$ correct agents under the assumption of at most $f$ faulty agents \cite{dolev:2000}. In dynamic epistemic logics it relates to {\em lying agents} and {\em bluffing agents} \cite{hvd.lying:2014} (somewhat corresponding to malicious respectively byzantine agents in distributed computing, as lying agents say what they know to be false with the intention to deceive, whereas bluffing agents say whatever they like unrelated to whether it is true, although this could also be with the intention to deceive).

\begin{quote} {\em 
We illustrate lying for, again, three children Anne, Bill and Cath. 

First, suppose only Anne is muddy, and suppose that Anne, incorrectly, does not step forward in the first round. Then Bill and Cath will mistakenly both step forward in the second round, as they incorrectly believe that Anne and Bill, respectively Anne and Cath, are muddy. 

Second, suppose Anne and Bill are muddy, and suppose Anne incorrectly steps forward in the first round. Then Bill incorrectly (but honestly) \emph{believes} that he is clean and steps forward in the second round. Whereas Cath \emph{knows} that Anne is incorrect: Cath sees two muddy children, so she knows that noone can know in the first round whether he or she is muddy.}
\end{quote}

\paragraph*{Muddy children with cleaning}

Yet further complications, not related in the published records of the dynasty of Henriettas, emerge when the value of factual propositions may change during a protocol's execution. Let us explain this by muddy children {\em with cleaning} \cite{hvdetal.puzzle:2015}. We should say though, that it is hard to imagine such factual change for unfaithful wives or husbands.

\begin{quote} {\em 
Suppose Anne and Bill are muddy and Cath is clean. After father has announced that at least one child is muddy, he takes a bucket of water and empties it over Anne (she knows he has a weird sense of humour sometimes, to which her mother entirely agrees, but, well, what can you say, he's her father). Father now, as ever, starts to repeat his request that those who know whether they are muddy step forward. In the first round, not surprisingly, Anne steps forward. Everyone expects this to happen, as it has become common knowledge that she is clean. In the second round, Bill will now not step forward, as Anne's action no longer means she was only seeing clean children. Bill cannot rule out that Anne would have stepped forward anyway, but also not that she would have waited.  We are stuck in this case. Father can repeat his request forever, and dinner will not be served.

However, let now Anne be clean whereas Bill and Cath are muddy. Father cleaning Anne \emph{seems} meaningless, as she was already clean. But appearances can be deceptive! Anne will still step forward in the first round. However, in the second round Bill and Cath will now, after all, step forward, and --- interestingly --- Anne learns from their behaviour that she must have been originally clean!}
\end{quote}

\paragraph*{Transfinite ordinals}

Finally a variation that is a bit harder to fit into three muddy children Anne, Bill and Cath. Parikh proposed a variation of {\em Consecutive Numbers} that requires a transfinite number of pairs of ignorance announcements
\begin{itemize}
\item Anne: ``I don't know your number.''
\item Bill: ``I don't know your number.''
\end{itemize}
before eventually first Anne knows Bill's and then Bill knows Anne's number, just as before \cite[p.\ 488]{parikh:1992}. 

\begin{figure}[h] \center
 \includegraphics[width=7cm]{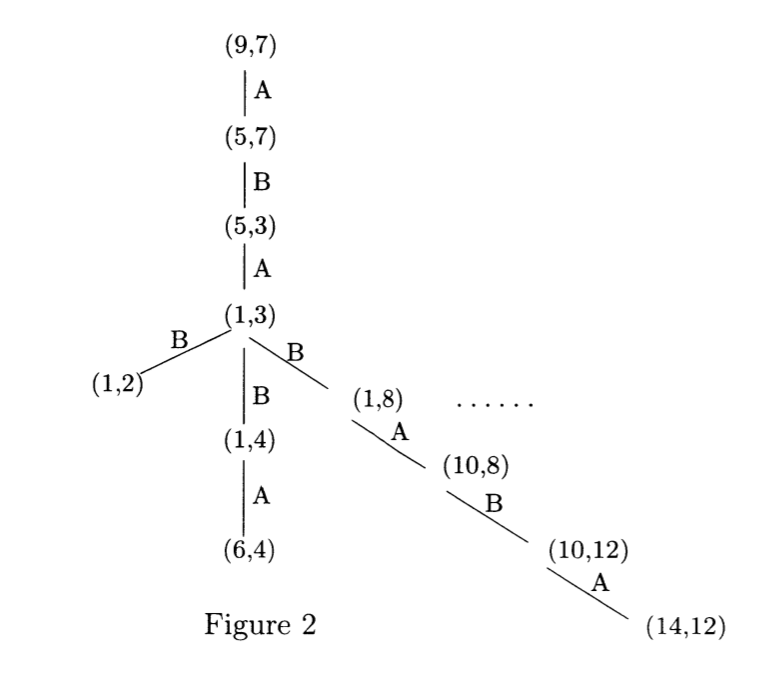}
\caption{An epistemic riddle with transfinitely many ignorance announcements \cite{parikh:1992}}
\label{figure.rohit}
\end{figure}

We can easily read the solution in the structure depicted in Figure~\ref{figure.rohit}, representing a domain of number pairs where an $A$-labelled link denotes indistinguishability for Anne and a $B$-labelled link indistinguishability for $B$ (and where, we cheat a bit, the first argument $i$ of pair $(i,j)$ is the number of $B$ and the second argument is that of $A$; in fact, Parikh uses the version where numbers are whispered to agents, not the version where numbers are on the forehead of --- other --- agents); the $\dots$ on the right of the figure summarizes an infinity of branches of length $2^k$ (for $k \geq 0$) rooted in $(1,3)$ where, to start with, $B$ is uncertain between $(1,3)$ and $(1,2^{k+1})$, and then as exemplified in the branches of length $2$ and $4$; whereas the branch with odd numbers should be an infinite one, which is not clear from the figure (there are missing $\dots$ so to speak --- it is easy to see this as a consequence of some misunderstanding in final typesetting). Note that the result of two ignorance announcements is frustatingly isomorphic to the initial model. But after $\omega$ such pairs of announcements all branches are gone!\footnote{Similar updates occur in \cite{agotnesetal.truelie:2017} illustrating that announcing lies can make them true.}

\section{M\"utzen: a novel hat puzzle with self-reference} \label{section.muetzen}

The Berlin Mathematics Research Center MATH+ annually presents an Advent calender with mathematical riddles, of which the solutions are subsequently given after Christmas. In the 2013 calender problem number 10, by Gerhard Woeginger, is called M\"utzen (`Hats') \cite{woegingermuetzen:2013}. It is a variation of muddy children and hat puzzles with many hats and many announcements, in a Christmas setting where Santa informs his Little Helpers (`Wichtel', Gnomes) of the solution requirement. It is accompanied by a beautiful illustration implicitly challenging the reader to determine the famous mathematicians depicted there as gnomes with coloured hats (Ren\'e Descartes, Leonhard Euler and Kurt G\"odel are among them). M\"utzen is given above in English translation. For the German original see \cite{woegingermuetzen:2013}, or see \cite{hvd.tark:2025} that includes a detailed analysis and formalization (sketched at the end of this section).\footnote{Everything about this puzzle is perfect. The  German original reads far better than my English translation. Take the many different colours, for example the buttercup yellow, kingcup yellow, primrose yellow and sunflower yellow hats. In German these are ``butterblumengelben, dotterblumengelben, schl\"usselblumengelben und sonnenblumengelben M\"utzen''. (Do not try playing Scrabble with German friends.) It is a perfect rhythm. And for any youngster roaming the fields and river banks when growing up, these are very recognizable and entirely different colours; kingcups and buttercups are fairly similar but kingcups are always a bit darker, whereas primroses have the lightest yellow you can imagine. And sunflowers are so dark yellow that they can be almost orange. These sweet memories of youth also come back with this puzzle!}

\medskip

{\small \em  
Santa Claus invited 126 smartgnomes to a cozy afternoon with coffee and cake. When the gnomes enter the hall, each of them gets a new gnome hat placed on their head from behind. This happens at lightning speed, so that none of them can see the colour of their own hat. Santa opens the meeting with a short speech.
\begin{quote}
My dear smartgnomes! We want to start this afternoon with a little brain teaser. None of you knows the colour of your own hat, and each of you can see the hats of all the other 125 gnomes. The aim of this game is to find out the colour of your own hat as quickly as possible and through pure thinking. I will now ring my big Christmas bell every five minutes. Once someone has found out their own hat colour, he must immediately leave the hall at the next ring of the bell. In the next room he is then served a cup of coffee and a large piece of Sachertorte.
\end{quote}
Santa is just about to ring the bell when gnome Atto comes up with an important question:
\begin{quote} Is it really possible for each of us to determine the colour of our hat through logical thinking? For example, if each of us had a different colour of hat, then no one would be able to figure out what colour it is by mere deduction. Then we couldn't win the game!
\end{quote}
Santa answers him a little gruffly:
\begin{quote}
Would, could, were!!! Of course, each of you can win this game! I chose the hat colours very carefully so that each of you can actually determine their colour through thinking during the game.
\end{quote}
And then the gnomes start thinking. And Santa starts ringing the bell.
\begin{itemize}
\item At the first ring of the bell, Atto and nine other gnomes leave the hall. \vspace{-.3cm}
\item At the second ring, all the gnomes walk out of the hall wearing buttercup yellow, kingcup yellow, primrose yellow and sunflower yellow hats. \vspace{-.3cm}
\item At the third ring all the gnomes leave with crimson hats, at the fourth ring all with cactus green hats, at the fifth ring all with aquamarine blue hats, at the sixth ring all with orange hats, at the seventh ring all with amber brown hats and at the eighth ring of the bell all leave with shell-gray hats. \vspace{-.3cm}
\item At the ninth, tenth, eleventh and twelfth rings no one leaves the hall. \vspace{-.3cm}
\item At the thirteenth ring, all the gnomes leave with blossom-white hats and all the gnomes with ebony-black hats.
\end{itemize}
And so it goes on. At the Nth ring of Santa, the last group of gnomes finally leaves the hall. Santa rang a total of seven times without anyone leaving the hall (and of these seven times, the ninth, tenth, eleventh and twelfth rings are already counted). Our question now is: What is N? Possible answers:
\[\begin{array}{|ll|ll|ll|ll|ll}
1. & N = 17 \quad & 3. & N = 19 \quad & 5. & N = 21 \quad & 7. & N = 23 \quad & 9. & N = 25 \quad \\
2. & N = 18 & 4. & N = 20 & 6. & N = 22 & 8. & N = 24 & 10. & N = 26 \qquad \hfill \cite{woegingermuetzen:2013}
\end{array}\]
}

\medskip


In the riddle's description we can easily see the pattern also followed in Muddy Children. First, the initial model is described $126$ gnomes only seeing the colours of others' hats. However, unlike Muddy Children, with declared `colours' black (muddy) and white (clean), nothing is known about these colours. Second, Santa makes a, in this case, curious initial announcement. It will play the role of father's initial announcement that at least one child is muddy. We call this announcement \solvable.
\begin{quote} {\em I chose the hat colours very carefully so that each of you can actually determine their colour through thinking during the game.} \hfill $(\solvable)$
\end{quote}
Third, the bell starts ringing, where at each ring those leave who know their colour, until finally all are having tea and cake. This is just as children stepping forward who know whether they are muddy, only in Muddy Children there are merely three stages, first nothing happens for many rounds, then all the muddy step forward, and then the remaining clean step forward. Whereas now there are many more stages.


The information content of \solvable\ is much harder to grasp than that of father's announcement that at least one child is muddy. The latter can be easily verified, unlike the former. Atto gives an important hint when he says to Santa ``For example, if each of us had a different colour of hat, then no one would be able to figure out what colour it is''. Let us be precise about Atto's observed inability to solve the riddle. 

If all gnomes wear a different unique colour, any gnome $i$ wearing colour $c$ also considers it possible that its hat has colour $c'$ for some colour $c' \neq c$ that is also not worn by any of the other gnomes. Therefore no gnome knows the colour of its hat and therefore no one will leave the room when the bell rings. And also not in the next iteration. Not ever. The later distribution of hats reveals that Atto cannot have seen $125$ all different colours. But this does not matter: as long as any gnome considers it possible that its own colour is unique, it considers it possible that this is any of two unseen colours and will therefore not leave the room when the bell rings. Therefore, it cannot have a unique colour, and therefore no gnome can have a unique colour, and therefore of each colour occurring there must be at least two hats. So that all the colours it sees, are all the colours there are. This demonstrates that the announcement \solvable\ is {\em at least as} informative as that of the announcement: 
\begin{quote} {\em For each coloured hat worn by a gnome there must be another gnome wearing a hat with the same colour. \hfill $(\solvable')$} \end{quote} We call that announcement $\solvable'$. This is indeed the spark that gets the induction going, so that finally all can determine their colour, and the problem can be solved. Intuitively, this  also demonstrates that \solvable\ is {\em at most as} informative as $\solvable'$, because what would otherwise justify ruling out even more colour distributions? Formally, it is the greatest fixpoint of the self-referential statement \solvable, which requires a bit more work to show \cite{hvd.tark:2025}. But therefore, \solvable\ and $\solvable'$ have the same information content. One can then proceed to solve the riddle just as all other muddy children and hat problems, using $\solvable'$ instead of \solvable.

Let us sketch how the iterations proceed that solve the riddle:

At the first iteration all gnomes seeing only one gnome with a hat of a particular colour conclude that they must also have that colour, and leave the room. As there were $10$ of those (Atto and nine others), that must concern five times two gnomes wearing some mutually distinct colours. At the second iteration all (only) seeing two gnomes with a hat of a particular colour, leave. As there are four shades of yellow, that must concern four times three is twelve gnomes. And so on. 

If in iteration $n$ no one sees $n$ gnomes of some colour, no one leaves the room. However, rather interesting I'd say, this is on condition, fulfilled in the riddle, that different colours can then still be seen. For example, let there be only $12$ gnomes, with $2$ white and $2$ black and $8$ rose hats. Then at ring $1$ the gnomes with white and black hats leave, but already at ring $2$ the gnomes with rose hats leave. \emph{It does not take more rounds.} But if there had been $10$ more pink and $10$ more yellow hatted gnomes, there would have been no such shortening of rounds for the $8$ rose hatted gnomes. Because then: at ring $1$ white and black leave, at rings $2$ to $6$ noone leaves, at ring $7$ the gnomes with rose hats leave, at ring $8$ noone leaves, and at ring $9$ the gnomes with pink and yellow hats leave. 

In the M\"utzen riddle shortening of rounds does not play a role, as there are two groups of the same maximum size that step forward in the last round. I do not know if this was by design, in order to avoid this extra modelling complication. It is a great pity I can no longer ask Gerhard Woeginger.

\begin{quote} {\em
With all these explanations, can you, reader, now solve M\"utzen? What is the correct answer from the ten given options?}
\end{quote}

\paragraph*{Formalizing M\"utzen in epistemic logic}
We succinctly describe the issues involving the formalization of M\"utzen given in \cite{hvd.tark:2025}.

With the reduction of the initial announcement $\solvable$ to $\solvable'$ we can formalize the M\"utzen riddle just as the Muddy Children Puzzle in public announcement logic (and indeed, modulo syntactic sugar, just as Plaza already did in his original publication \cite{plaza:1989}), i.e.\ a propositional modal logic with epistemic modalities for knowledge for individual agents and dynamic modalities for announcements, inducing updates of epistemic models consisting of alternative distributions of mud over children's foreheads and, for each agent, binary indistinguishability relations between such distributions. In other words: just as in Figure~\ref{figure.rohit}. Already one can justify a simplification here: instead of arbitrarily many, that is infinitely many, colours, it suffices to have one more colour than the number of $126$ gnomes (guaranteeing that each gnome is uncertain between at least two unseen colours for its own hat), so that conjunctions and such over the set of all distributions of $127$ colours over $126$ gnomes remain finite. This simplifies the formalization in a {\em propositional} modal logic. An alternative formalization, suggested by Barteld Kooi, where one merely formalizes whether gnomes have different colours, does not need such a simplication.

Formalizing $\solvable$ is a bit harder. We recall the self-referential nature of the announcement $\solvable$: given an epistemic model consisting of the domain of colour distributions over gnomes and indistinguishability between them, it describes a restriction of that domain such that in that restriction gnomes will always eventually learn their colour. This is a \emph{fixpoint}, and we therefore need a fixpoint extension of epistemic logic, a {\em modal $\mu$-calculus} \cite{Kozen82} (with modalities for knowledge). 

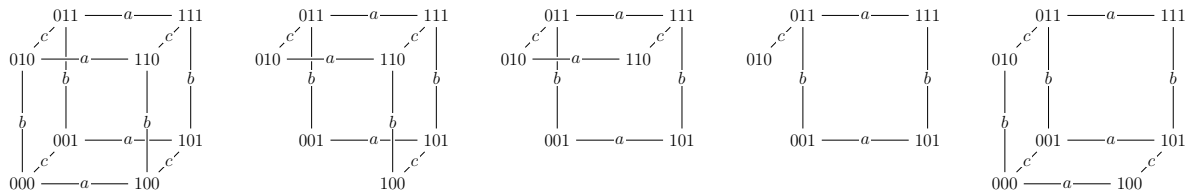
\begin{figure}[h]
\scalebox{.55}{
\begin{tikzpicture}[z=0.35cm]
\node (000) at (0,0,0) {$000$};
\node (001) at (0,0,3) {$001$};
\node (010) at (0,3,0) {$010$};
\node (011) at (0,3,3) {$011$};
\node (100) at (3,0,0) {$100$};
\node (101) at (3,0,3) {$101$};
\node (110) at (3,3,0) {$110$};
\node (111) at (3,3,3) {$111$};
\draw (000) -- node[fill=white,inner sep=1pt] {$a$} (100);
\draw (001) -- node[fill=white,inner sep=1pt] {$a$} (101);
\draw (011) -- node[fill=white,inner sep=1pt] {$a$} (111);
\draw (000) -- node[fill=white,inner sep=1pt] {$b$} (010);
\draw (001) -- node[fill=white,inner sep=1pt] {$b$} (011);
\draw (101) -- node[fill=white,inner sep=1pt] {$b$} (111);
\draw (000) -- node[fill=white,inner sep=1pt] {$c$} (001);
\draw (010) -- node[fill=white,inner sep=1pt] {$c$} (011);
\draw (100) -- node[fill=white,inner sep=1pt] {$c$} (101);
\draw (110) -- node[fill=white,inner sep=1pt] {$c$} (111);
\draw[draw=white,double=black,very thick] (010) -- node[fill=white,inner sep=1pt] {$a$} (110);
\draw[draw=white,double=black,very thick] (100) -- node[fill=white,inner sep=1pt] {$b$} (110);
\end{tikzpicture}
\qquad
\begin{tikzpicture}[z=0.35cm]
\node (000) at (0,0,0) {\color{white}$000$};
\node (001) at (0,0,3) {$001$};
\node (010) at (0,3,0) {$010$};
\node (011) at (0,3,3) {$011$};
\node (100) at (3,0,0) {$100$};
\node (101) at (3,0,3) {$101$};
\node (110) at (3,3,0) {$110$};
\node (111) at (3,3,3) {$111$};
%
\draw (001) -- node[fill=white,inner sep=1pt] {$a$} (101);
\draw (011) -- node[fill=white,inner sep=1pt] {$a$} (111);
%
\draw (001) -- node[fill=white,inner sep=1pt] {$b$} (011);
\draw (101) -- node[fill=white,inner sep=1pt] {$b$} (111);
%
\draw (010) -- node[fill=white,inner sep=1pt] {$c$} (011);
\draw (100) -- node[fill=white,inner sep=1pt] {$c$} (101);
\draw (110) -- node[fill=white,inner sep=1pt] {$c$} (111);
\draw[draw=white,double=black,very thick] (010) -- node[fill=white,inner sep=1pt] {$a$} (110);
\draw[draw=white,double=black,very thick] (100) -- node[fill=white,inner sep=1pt] {$b$} (110);
\end{tikzpicture} 
\qquad
\begin{tikzpicture}[z=0.35cm]
\node (000) at (0,0,0) {\color{white}$000$};
\node (001) at (0,0,3) {$001$};
\node (010) at (0,3,0) {$010$};
\node (011) at (0,3,3) {$011$};
\node (101) at (3,0,3) {$101$};
\node (110) at (3,3,0) {$110$};
\node (111) at (3,3,3) {$111$};
%
\draw (001) -- node[fill=white,inner sep=1pt] {$a$} (101);
\draw (011) -- node[fill=white,inner sep=1pt] {$a$} (111);
%
\draw (001) -- node[fill=white,inner sep=1pt] {$b$} (011);
\draw (101) -- node[fill=white,inner sep=1pt] {$b$} (111);
%
\draw (010) -- node[fill=white,inner sep=1pt] {$c$} (011);
\draw (110) -- node[fill=white,inner sep=1pt] {$c$} (111);
\draw[draw=white,double=black,very thick] (010) -- node[fill=white,inner sep=1pt] {$a$} (110);
%
\end{tikzpicture} 
\qquad
\begin{tikzpicture}[z=0.35cm]
\node (000) at (0,0,0) {\color{white}$000$};
\node (001) at (0,0,3) {$001$};
\node (010) at (0,3,0) {$010$};
\node (011) at (0,3,3) {$011$};
\node (101) at (3,0,3) {$101$};
\node (111) at (3,3,3) {$111$};
%
\draw (001) -- node[fill=white,inner sep=1pt] {$a$} (101);
\draw (011) -- node[fill=white,inner sep=1pt] {$a$} (111);
%
\draw (001) -- node[fill=white,inner sep=1pt] {$b$} (011);
\draw (101) -- node[fill=white,inner sep=1pt] {$b$} (111);
%
\draw (010) -- node[fill=white,inner sep=1pt] {$c$} (011);
%
%
\end{tikzpicture} 
\qquad
\begin{tikzpicture}[z=0.35cm]
\node (000) at (0,0,0) {$000$};
\node (001) at (0,0,3) {$001$};
\node (010) at (0,3,0) {$010$};
\node (011) at (0,3,3) {$011$};
\node (100) at (3,0,0) {$100$};
\node (101) at (3,0,3) {$101$};
\node (111) at (3,3,3) {$111$};
\draw (000) -- node[fill=white,inner sep=1pt] {$a$} (100);
\draw (001) -- node[fill=white,inner sep=1pt] {$a$} (101);
\draw (011) -- node[fill=white,inner sep=1pt] {$a$} (111);
\draw (000) -- node[fill=white,inner sep=1pt] {$b$} (010);
\draw (001) -- node[fill=white,inner sep=1pt] {$b$} (011);
\draw (101) -- node[fill=white,inner sep=1pt] {$b$} (111);
\draw (000) -- node[fill=white,inner sep=1pt] {$c$} (001);
\draw (010) -- node[fill=white,inner sep=1pt] {$c$} (011);
\draw (100) -- node[fill=white,inner sep=1pt] {$c$} (101);
%
%
\end{tikzpicture}
}
\caption{Solvable and unsolvable restrictions of initial uncertainty in (three) Muddy Children. From left to right: initial uncertainty, solvable, unsolvable, solvable. All restrictions to seven are solvable, rightmost another one.}
\label{fig.differentrestr}
\end{figure}

To determine the right (fixpoint) subdomain of the given initial model of colour distributions, we cannot `simply' reduce the entire domain, distribution by distribution, until we are there, or expand the actual colour distribution similarly until we are there. These expansions or restrictions have to satisfy the puzzle constraints (solvability) and thus contains jumps adding or deleting sets of multiple distributions at the same time. Consider --- as ever --- three muddy children, and Figure~\ref{fig.differentrestr} depicting some restrictions of the initial epistemic model of uncertainty with eight distributions of mud (where $100$ means `Anne muddy, Bill clean, Cath clean', etcetera). The standard solution, the restriction to seven, without all three children clean, is sufficient to solve the riddle. But not the depicted restriction to six distributions. It is easy to see that Anne remains forever uncertain between $011$ and $111$: at father's first request to step forward, Bill steps forward in $010$ and $110$ and Cath steps forward in $001$ and $101$; at father's second request Bill and Cath step forward in $011$ and $111$; no further requests will remove Anne's uncertainty. But remove one more distribution, e.g.\ $110$, and it is solvable again. In other words, only some restrictions satisfy the puzzle constraints.

But there is more. The announcement in M\"utzen is colour-blind: no particular colours are mentioned. This implies that we are not looking for a domain restriction as in Muddy Children. Because any restriction of the initial (leftmost) epistemic model in Figure~\ref{fig.differentrestr} to seven distributions is sufficient to eventually have all three children learn whether they are muddy. Another example is the rightmost restriction in Figure~\ref{fig.differentrestr}. It is the result of father announcing ``It is not the case that Anne and Bill are muddy and Cath is clean'' and the further reduction in uncertainty after father's request is analogous to the standard one. If the actual distribution was $001$, this becomes (again) known after three requests. But these two seven-distribution restrictions cannot be transformed into one another by a permutation of clean into muddy and muddy into clean. We then get yet two other seven-distribution restrictions. Colour-blindness in M\"utzen means that we are looking for restrictions that are {\em invariant under any permutation of colours over hats}. If there had been four colours for the three children instead of two (clean and muddy), the only colour-blind restriction would consist of the four distributions where they have all the same colour. In all of these cases where all only see one colour, they can conclude it must be their own colour too. All will immediately step forward / have cake.

A final complication is to prove that fixpoints exist. For modal logical language extensions with fixpoint operators $\mu x.\phi$ (least fixpoint) and $\nu x .\phi$ (greatest fixpoint), the existence of fixpoints is guaranteed if the occurrences of $x$ in $\phi$ are positive, that is, bound by an even number of negations in a (negation) normal form. Unfortunately, for further language extensions with also dynamic modalities for announcements no such inductively defined language fragments can be given.\footnote{After all a formula! Consider $\dia{q}K p$ --- after the true announcement of $q$, $p$ is known --- and $\dia{q} \M p$ --- after the true announcement of $q$, $p$ is considered possible. These formulas are logically equivalent to $q \et \M(q \et p)$ respectively $q \et K(\neg q \vel p)$, where $q$ only occurs positively in the first, but not in the second.} One therefore has to prove existence of fixpoints `the hard way', showing monotony.

Fixpoint semantics have been investigated for epistemic modal logics in \cite{HalpernM86,parikh:1992,jfaketal.mu:2008,BaltagBF22,yanjunetal:2025}. These more or less straightforward adapt the modal $\mu$-calculus \cite{Kozen82}. The issues when combining announcements and fixpoints are well-known \cite{jfaketal.mu:2008,BaltagBF22,yanjunetal:2025}, and in \cite{hvd.tark:2025} only somewhat further developed but mainly applied to M\"utzen: the formula $\solvable$ is then a greatest fixpoint binding a formula containing a least fixpoint \cite[p.\ 443]{hvd.tark:2025}.

\paragraph*{Acknowledgements} Detailed acknowledgements will be added later.

\bibliographystyle{plain}
\bibliography{biblio2026}

\end{document}